\documentclass{article} 
\usepackage{iclr2027_conference,times}

\usepackage{amsmath,amsfonts,bm}

\def\eqref#1{equation~\ref{#1}}

\def\1{\bm{1}}

\DeclareMathAlphabet{\mathsfit}{\encodingdefault}{\sfdefault}{m}{sl}
\SetMathAlphabet{\mathsfit}{bold}{\encodingdefault}{\sfdefault}{bx}{n}

\newcommand{\R}{\mathbb{R}}

\newcommand{\softmax}{\mathrm{softmax}}

\usepackage{hyperref}
\usepackage{url}
\usepackage{graphicx}
\usepackage{cancel}
\usepackage{svg}

\usepackage{amsthm}
\usepackage{xcolor}
\theoremstyle{remark}

\usepackage{tcolorbox}

\usepackage{hyperref}
\usepackage{url}
\usepackage{graphicx}
\usepackage{cancel}
\usepackage{svg}

\usepackage{amsthm}
\usepackage{xcolor}

\definecolor{referenceblue}{rgb}{0.39,0.58,0.93}
\hypersetup{
    colorlinks=true,
    linkcolor=referenceblue,
    citecolor=referenceblue,
    urlcolor=referenceblue
}

\newcommand{\one}{\mathbf{1}}
\newcommand{\norm}[1]{\left\lVert #1\right\rVert}
\newcommand{\abs}[1]{\left\lvert #1\right\rvert}
\newcommand{\inner}[2]{\left\langle #1,#2\right\rangle}
\newcommand{\varnorm}[1]{\left\lVert #1\right\rVert_{\mathrm{var}}}
\DeclareMathOperator{\osc}{osc}

\title{Taming Softmax at the Limits of Muon: A \\ Unified Approach through Hilbert Geometry}

\author{%
  \textbf{Aditya Somasundaram, Charles Guille-Escuret, Alexander Moreno} \\
  \textbf{Zhengzhong Liu, Eric Xing} \\[4pt]
  \normalfont Institute of Foundation Models, \\
  \normalfont Mohamed bin Zayed University of Artificial Intelligence (MBZUAI) \\
  \normalfont \texttt{aditya.somasundaram@mbzuai.ac.ae}
}

\iclrfinalcopy
\begin{document}

\maketitle
\lhead{Preprint. Under review.}

\begin{abstract}
Muon controls weight updates through the RMS movement of a linear layer's output.
This implicitly treats the linear output as a proxy for the full module, including the nonlinearity that follows.
This proxy can be misleading for softmax, where the RMS shift in activations does not faithfully measure the change in the output distribution.
We propose Hilbert distance as a natural metric for bounding changes in these output distributions.
Hilbert distance bounds changes in cross-entropy loss and limits probability redistribution.
For softmax, it takes the form of the variation seminorm of the preactivation perturbation, allowing probability constraints to be expressed as parameter constraints.
Formulating the steepest descent problem with a Hilbert distance constraint brings two recently proposed methods under a unified mathematical framework.
For the LM head, solving a row-wise approximation to the resulting diameter-constrained problem leads us to RowNorm \citep{pethick2025trainingdeeplearningmodels}.
For scaled dot-product attention, the same principle yields the first-order query--key steepest descent problem underlying Compositional Muon \citep{keigwin2026compositional}.
These connections motivate using Hilbert distance as a guiding principle for optimizer design in softmax modules.
\end{abstract}

\section{Introduction}

An important consideration for stable training is the extent to which a single update changes the network's intermediate representations.
Each intermediate representation is both the output of one layer and the input to the next.
This dependence is especially important in deep sequential models, where changing one representation alters the inputs processed by every layer that follows.
Without proper control, these representation shifts can accumulate across layers and destabilize training.
Muon formalizes this idea for linear layers by controlling the worst-case RMS movement of the layer's output over inputs of bounded size \citep{jordan2024muon,bernstein2025deriving}.
This requirement leads to a suitably scaled spectral norm constraint on the weight update.


This spectral constraint is not arbitrary: it follows from choosing RMS to measure the functional change produced by the update.
More broadly, it illustrates how the geometry of an optimizer depends on the notion of change assigned to a module.
The Modula project and Scion generalize this perspective by advocating update geometries that reflect the function of each module \citep{modula-docs,pethick2025trainingdeeplearningmodels}.

Muon's standard update geometry, however, is derived for the linear map and does not explicitly adapt to architectural operations such as normalization and nonlinear activations.
As a result, Muon does not yet provide a single optimization rule for every parameter group in the model.
In standard Muon training recipes, the LM head, embeddings, and scalar or vector parameters are typically optimized with AdamW \citep{jordan2024muon}.
Large-scale Muon training can also require additional stabilization.
For example, Kimi K2 introduced MuonClip to address excessive attention-score growth by rescaling query and key weights after an update \citep{kimiteam2026kimik2openagentic}.

We focus on how optimizer design changes when a linear map is followed by softmax.
The linear layer's output is an intermediate state, and its RMS displacement does not faithfully measure the change in the probability distribution produced by softmax.
We instead directly constrain the change in the output probability distribution.
We propose using Hilbert's projective metric, or Hilbert distance, for this purpose.


Our contributions are threefold:
\begin{itemize}
    \item We formulate the steepest descent problem using Hilbert distance to constrain changes in softmax output distributions. This single formulation brings two seemingly independent optimization methods under a unified mathematical framework.
    \item For the LM head, we show that this probability constraint becomes a row diameter constraint. Its exact dual scales quadratically with vocabulary size. A tighter common-ball constraint instead leads to RowNorm in closed form \citep{pethick2025trainingdeeplearningmodels}.
    \item For standard attention, we show that this probability constraint produces a spectral constraint on the first-order change in the QK product, recovering the steepest descent problem posed in Compositional Muon \citep{keigwin2026compositional}.
\end{itemize}


Beyond unifying these recent methods, we provide a guiding principle for optimizer design for any neural network module that outputs a probability distribution.

We first expose the mismatch between RMS and softmax (Section~\ref{sec:muon-preliminaries}).
Next, we introduce Hilbert geometry and establish its suitability for measuring changes in softmax output distributions (Section~\ref{sec:hilbert}).
We subsequently formulate the corresponding steepest descent problems for the LM head and attention.
For the LM head, a tighter approximation leads to RowNorm in closed form.
For attention, our formulation recovers the QK steepest descent problem studied by Compositional Muon
(Sections~\ref{sec:lm-head} and~\ref{sec:compositional-muon}).

\section{Preliminaries on steepest descent}
\label{sec:muon-preliminaries}

Steepest descent chooses an update that gives the greatest first-order decrease in the loss while remaining small under a chosen measure of size.
For a minibatch gradient $G=\nabla_W\mathcal L$, an update $\Delta W$ changes the loss to first order by $\inner{G}{\Delta W}_F$.
Let $\mathcal D_W(\Delta W)$ measure the size of this update at the current weights $W$.
We consider
\begin{equation}
\min_{\Delta W} \inner{G}{\Delta W}_F
\qquad
\text{subject to}
\qquad
\mathcal D_W(\Delta W)\leq\rho,
\label{eq:steepest-descent}
\end{equation}
where $\rho>0$ sets the permitted update size.
We leave $\mathcal D_W$ unspecified for now, allowing its choice to reflect the module being optimized.




The importance of the underlying choice of metric cannot be overstated.
Choosing the Frobenius norm measures the update directly in parameter space and gives the SGD direction \citep{robbins1951stochastic}.
Muon takes a functional perspective instead: for a linear layer, it controls the worst-case change in the output produced by the update \citep{bernstein2025deriving}.
Measuring this change with RMS leads to a spectral norm constraint on the weight update, whose steepest descent solution gives the Muon direction.
The Modula project generalizes this perspective by advocating update geometries that reflect the function of each module \citep{modula-docs}.
Around this principle, Scion develops a systematic optimization framework, formulating updates as Linear Minimization Oracles under $\norm{\cdot}_{p\rightarrow q}$ operator norm constraints \citep{pethick2025trainingdeeplearningmodels}.
Here, the norms $\norm{\cdot}_p$ and $\norm{\cdot}_q$ describe the input and output geometries of the module.
The shared ethos of these works is to build optimizers whose geometry reflects the function of each module.

This functional perspective becomes more delicate when a linear layer is followed by a nonlinearity.
The linear output is then only an intermediate state in the computation of the full module.
Controlling the RMS movement of this intermediate state does not account for the geometry introduced by the subsequent nonlinearity.
For a softmax module, the linear output is the preactivation vector, which softmax maps to a probability distribution on the simplex.
We therefore ask whether RMS displacement of the preactivation faithfully measures changes in the output distribution.


\begin{figure}[t]
    \centering
    \includegraphics[width=\linewidth]{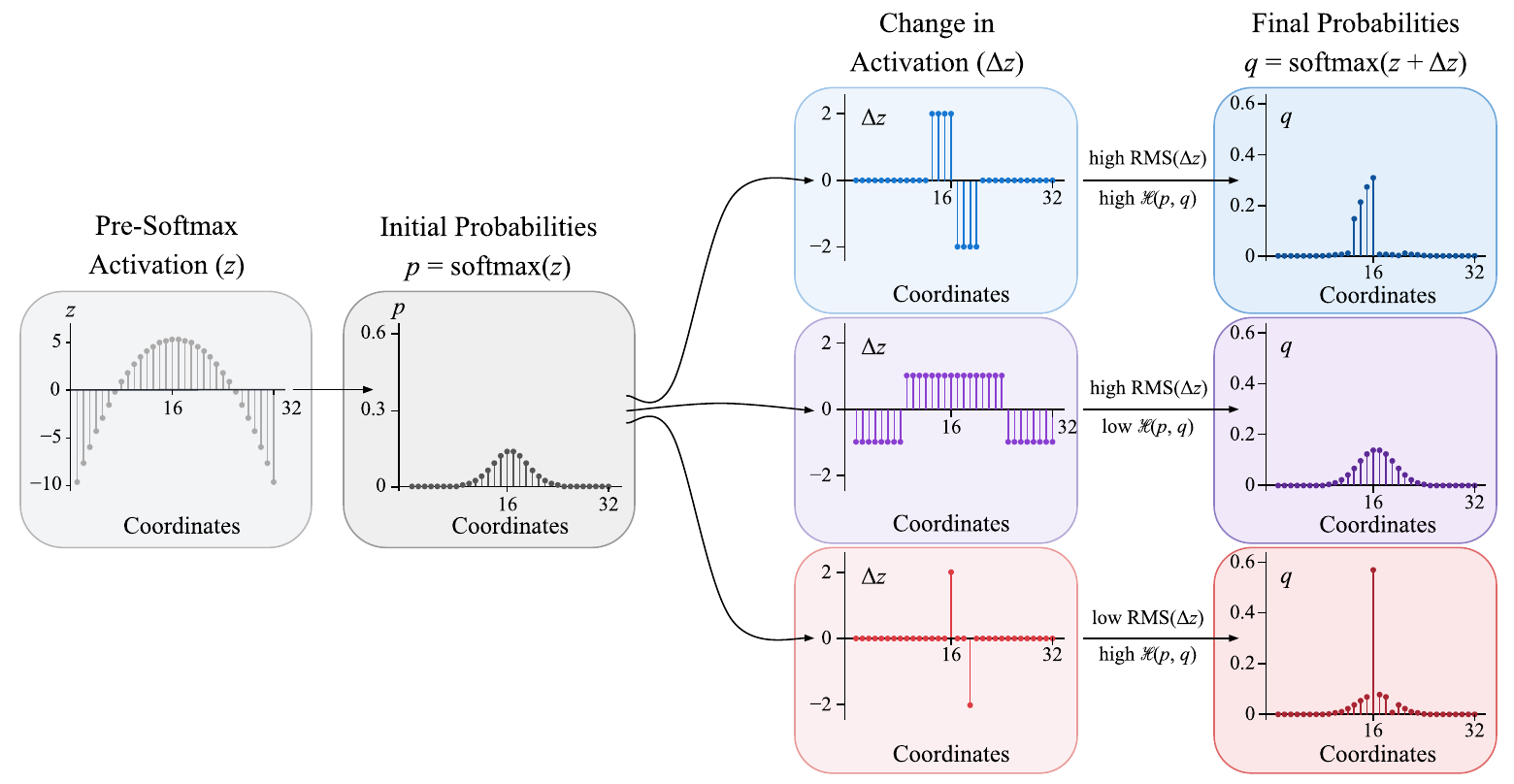}
    \caption{\textbf{A common RMS bound permits very different changes in softmax probabilities.}
All cases share the same centered $z\in\R^{32}$ and initial distribution $p=\softmax(z)$.
The centered perturbations produce $q=\softmax(z+\Delta z)$ and all satisfy $\mathrm{RMS}(\Delta z)\leq1$.
The pairs $(\mathrm{RMS}(\Delta z),\mathcal H(p,q))$ are \textcolor{blue}{$(1,4)$} for \textcolor{blue}{blue}, \textcolor{violet}{$(1,2)$} for \textcolor{violet}{purple}, and \textcolor{red}{$(1/2,4)$} for \textcolor{red}{red}, where $\mathcal H$ denotes Hilbert distance.
\textcolor{blue}{Blue} substantially reshapes the distribution, while \textcolor{violet}{purple} barely changes it.
 \textcolor{red}{Red} matches \textcolor{blue}{blue's} Hilbert distance at half its RMS.}
    \label{fig:rms-softmax-distributions}
\end{figure}

\subsection{Geometric mismatch between Muon and softmax}
We argue that Muon is 
inadequate for modules involving softmax. Muon's spectral geometry can allocate updates poorly for such modules. The distinction lies between the nature of RMS and softmax: RMS measures how much the preactivation vector moves on average, while softmax measures how much one coordinate \textit{gains} relative to \textit{others}.

\paragraph{A simple example.}
Consider an update where all the rows are equal, $\Delta W = \one b^\top$.
In this case, $\Delta z = \Delta W x = (b^\top x) \one$, a scaled ones vector.
Such a perturbation to the preactivation vector does not alter the probability distribution, despite incurring an arbitrarily large RMS norm.
This particular case is not problematic in practice because softmax modules receive row-centered gradients, causing Muon to consider only centered directions; see Section~\ref{sec:discussion}. It nevertheless illustrates a broader geometric mismatch between RMS bounds and softmax that can produce practical pathologies.

Figure~\ref{fig:rms-softmax-distributions} shows that this discrepancy persists under centered perturbations.
Unlike the preceding example, all three cases share the same initial distribution and use centered perturbations satisfying a common RMS bound.
The \textcolor{blue}{blue} and \textcolor{violet}{purple} updates have equal RMS, yet the \textcolor{blue}{blue} perturbation substantially reshapes the distribution while the \textcolor{violet}{purple} perturbation barely changes it.
The \textcolor{red}{red} perturbation has half their RMS but still causes a substantial probability redistribution.

This mismatch compels direct control of the output distribution. Designing an effective update rule for softmax modules requires choosing a suitable metric on the output distribution and formulating the induced steepest descent problem. This raises a fundamental question:

\begin{quote}
\emph{What is the appropriate metric for perturbations of the softmax probabilities?}
\end{quote}

\section{Hilbert geometry of softmax}
\label{sec:hilbert}





We turn to Hilbert's projective metric, or simply Hilbert distance  \citep{birkhoff1957extensions, de1993hilbert}.
For strictly positive vectors $x,y\in\R_{>0}^n$, it is defined by
\(
\mathcal H(x,y)
=
\log
\frac{\max_i x_i/y_i}{\min_j x_j/y_j}.
\)
We apply this metric to strictly positive probability distributions on the simplex
$\mathcal P=\{x\in\R_{>0}^n:\sum_i x_i=1\}$ \citep{Nielsen_2018, nielsen2023non}.
For $p,q\in\mathcal P$, Hilbert distance can equivalently be written as
\begin{equation}
\mathcal H(p,q)
=
\max_{i,j}
\left|
\log
\frac{q_i/q_j}{p_i/p_j}
\right|.
\label{eq:hilbert-definition}
\end{equation}
Thus, Hilbert distance measures the largest change in pairwise odds.

Hilbert's projective metric has several useful geometric properties on positive cones and probability simplices \citep{lemmens2012nonlinear, papadopoulos2014handbook}. Bounding the Hilbert distance directly controls the change in cross-entropy loss and limits probability redistribution in total variation. For softmax, the distance also admits a closed-form expression in terms of the preactivation update: it coincides with the range between the largest and smallest coordinates (the variation seminorm). Consequently, uniform shifts leave the distance invariant, mirroring the translation invariance of the softmax itself. Motivated by these geometric properties, we adopt the Hilbert distance to quantify changes in output distributions and formulate the resulting steepest descent problem.

\paragraph{Hilbert distance bounds pairwise odds ratios.}
Throughout, let $z\in\R^n$ denote the preactivation supplied to softmax, and let $p=\softmax(z)$ be the resulting output distribution.
A perturbation $\Delta z$ produces the updated distribution $q=\softmax(z+\Delta z)$.
By Equation~\ref{eq:hilbert-definition}, bounding Hilbert distance uniformly bounds the odds ratio for every pair of outcomes:
\[
\mathcal H(p,q)\leq\varepsilon
\quad\Longleftrightarrow\quad
e^{-\varepsilon}
\leq
\frac{q_i/q_j}{p_i/p_j}
\leq
e^\varepsilon
\qquad
\forall i,j.
\]
In particular, for softmax, Hilbert distance has the closed form
\begin{equation}
\mathcal H\bigl(\softmax(z),\softmax(z+\Delta z)\bigr)
=
\max_{i,j}\abs{\Delta z_i-\Delta z_j}
=
\max_i\Delta z_i-\min_i\Delta z_i
=
\varnorm{\Delta z}.
\label{eq:hilbert-softmax}
\end{equation}
The variation seminorm $\varnorm{\Delta z}$, also called the oscillation $\osc(\Delta z)$, is the difference between the largest and smallest entries of $\Delta z$.
It vanishes precisely when $\Delta z$ is constant, matching the translation invariance of softmax.

\paragraph{Hilbert distance bounds probability redistribution.}
Bounding pairwise odds ratios also limits the redistribution of probability mass.
We measure this redistribution using total variation,
\[
\mathrm{TV}(p,q)
=
\frac12\sum_i\abs{p_i-q_i}
=
\max_{A\subseteq\{1,\ldots,n\}}
\abs{\sum_{i\in A}(p_i-q_i)}.
\]
Thus, $\mathrm{TV}(p,q)\leq\delta$ guarantees that no set of outcomes can gain or lose more than $\delta$ probability mass.
Hilbert distance gives a sharp bound on total variation \citep[Theorem~5.1]{cohen2024hyperboliccontractivityhilbertmetric},
\[
\mathrm{TV}(p,q)
\leq
\tanh\!\left(\frac{\mathcal{H}(p,q)}4\right).
\]
The bound is sharp even when the preactivation perturbation is fixed.
For every $\Delta z\in\R^n$,
\[
\sup_{z\in\R^n}
\mathrm{TV}\bigl(\softmax(z),\softmax(z+\Delta z)\bigr)
=
\tanh\!\left(\frac{\varnorm{\Delta z}}4\right).
\]
The variation seminorm therefore determines the largest probability redistribution that a given preactivation update can produce across all starting distributions. Appendix~\ref{app:softmax-tv} proves this sharpness statement.

\paragraph{Hilbert distance bounds change in cross-entropy (CE) loss.}
Let $p$ and $q$ be the prediction distributions before and after an update.
For a target outcome $i$, the CE loss is $\ell_i(p) = -\log p_i$, so its change is $\Delta \ell_i = \ell_i(q) - \ell_i(p) = \log \frac{p_i}{q_i}$.
Hilbert distance is the range of these changes, $\mathcal{H}(p, q) = \max_i \Delta \ell_i - \min_i \Delta \ell_i$.
It therefore measures how differently the update affects the loss across possible targets.
Moreover, $\sum_i p_i = \sum_i q_i = 1$ implies $\min_i \frac{p_i}{q_i} \leq 1 \leq \max_i \frac{p_i}{q_i}$.
Taking logarithms gives $\min_i \Delta \ell_i \leq 0 \leq \max_i \Delta \ell_i$.
Consequently, $\mathcal H(p,q)\leq\varepsilon$ guarantees that the change in CE loss is at most $\varepsilon$,
\[
\abs{\Delta \ell_i} \leq \max_j \Delta \ell_j - \min_j \Delta \ell_j \leq \mathcal{H}(p, q) \quad \forall i.
\]



\paragraph{Steepest descent problem.}
Together, these results show that Hilbert distance controls pairwise odds, probability redistribution, and changes in cross-entropy loss.
Having established these properties, we now formulate the steepest descent problem using Hilbert distance to measure the size of an update.
We return to Equation~\ref{eq:steepest-descent} and express the update constraint in terms of this distance.
Let $W$ denote the module's parameters and let $p_W(x)$ be the output distribution produced at input $x$.
An update $\Delta W$ changes this distribution from $p_W(x)$ to $p_{W+\Delta W}(x)$.
Requiring these distributions to remain within Hilbert distance $\varepsilon$ for every input in a prescribed set $\mathcal X$ gives
\begin{equation}
\min_{\Delta W}
\inner{\nabla_W\mathcal L}{\Delta W}_F
\qquad
\text{subject to}
\qquad
\sup_{x\in\mathcal X}
\mathcal H\bigl(p_W(x),p_{W+\Delta W}(x)\bigr)
\leq\varepsilon.
\label{eq:hilbert-steepest-descent}
\end{equation}

\section{The LM Head}
\label{sec:lm-head}
For the LM head, bounding Hilbert distance over the hidden state space is equivalent to bounding the diameter of the update rows.
The resulting steepest descent problem has an exact dual whose size grows quadratically with the vocabulary.
A tighter common-ball approximation avoids this coupling and leads to RowNorm in closed form.

We first specialize Equation~\ref{eq:hilbert-steepest-descent} to the LM head.
For a hidden state $h\in\R^d$, the LM head produces the logit vector $z=Wh\in\R^V$ and the token distribution $p=\softmax(z)$.
We consider an untied head without bias, trained with CE and no other loss term depending on $W$.
Throughout this section, we compare the head before and after an update at the same hidden state.

\paragraph{Hilbert distance induces a row diameter constraint.}
Let $\Delta W$ be a weight update with rows $\Delta w_i^\top$.
The update changes the logit for token $i$ by $\Delta w_i^\top h$.
For two tokens $i$ and $j$, their logit difference therefore changes by $(\Delta w_i-\Delta w_j)^\top h$.
Equation~\ref{eq:hilbert-softmax} gives
\[
\mathcal{H}\bigl(p_W(h),p_{W+\Delta W}(h)\bigr)
=\varnorm{\Delta Wh}
=\max_{i,j}\abs{(\Delta w_i-\Delta w_j)^\top h}.
\]
For a fixed pair of rows, the largest value of this inner product over $\norm{h}_2\leq R$ is $R\norm{\Delta w_i-\Delta w_j}_2$.
Taking the largest value over all row pairs gives
\begin{equation}
\sup_{\norm{h}_2\leq R}
\mathcal{H}\bigl(p_W(h),p_{W+\Delta W}(h)\bigr)
=R\max_{i,j}\norm{\Delta w_i-\Delta w_j}_2
=R D(\Delta W).
\label{eq:head-hilbert-diameter}
\end{equation}
The maximum pairwise distance is precisely the diameter $D(\Delta W)$ of the row cloud $\{\Delta w_1,\ldots,\Delta w_V\}$ \citep[Definition~3.9]{rudin1976principles}.
For this input ball, the general constraint therefore specializes to $\mathcal D_W(\Delta W)=R D(\Delta W)$.
Choosing $R=\sqrt d$ gives the RMS unit ball.
Thus, requiring a Hilbert distance of at most $\varepsilon$ throughout this ball is equivalent to imposing $D(\Delta W)\leq\rho_{\mathrm{head}}$, where $\rho_{\mathrm{head}}=\varepsilon/R$.

This single constraint on the update rows controls both the predicted distribution and the cross-entropy loss.
By Section~\ref{sec:hilbert}, no group of tokens can gain or lose more than $\tanh(\varepsilon/4)$ probability mass, and the CE loss changes by at most $\varepsilon$ for any target token.

\paragraph{Steepest descent under the row diameter constraint.}
We seek the update with the greatest first-order loss decrease under the row diameter bound.
Let $G=\nabla_W\mathcal{L}$ have rows $g_i^\top$.
Using the row diameter constraint in Equation~\ref{eq:head-hilbert-diameter}, we obtain
\begin{equation}
\min_{\Delta W} \inner{G}{\Delta W}_F
\qquad
\text{subject to}
\qquad
D(\Delta W)\leq\rho_{\mathrm{head}}.
\label{eq:head-diameter-descent}
\end{equation}
Since the loss depends on $W$ only through the softmax output, its gradient satisfies $\one^\top G=0$.
Adding the same vector to every update row produces a common shift of the logits and therefore leaves both the output distribution and the loss unchanged.
Because pairwise row differences are also unchanged, the diameter constraint places no restriction on the common translation of the update rows.

Appendix~\ref{app:head-duality} gives an exact dual characterization of this problem.
The dual is a vector-valued optimal transport problem \citep{ciosmak2021optimal}.
In our setting, this transport problem introduces a $d$-dimensional flow variable for each of the $\binom V2$ token pairs.
This quadratic growth with vocabulary size makes solving the flow problem computationally expensive.
We proceed on a quest to find a cheaper approximation.

\paragraph{A common-ball approximation leads to RowNorm.}
We ask for a tighter constraint, requiring all update rows to lie in a common closed Euclidean ball $\mathcal B(\zeta,\rho_{\mathrm{head}}/2)$.
With this restriction, Equation~\ref{eq:head-diameter-descent} becomes
\begin{equation}
\min_{\Delta W,\zeta} \inner{G}{\Delta W}_F
\qquad
\text{subject to}
\qquad
\Delta w_i\in\mathcal B(\zeta,\rho_{\mathrm{head}}/2)
\quad\forall i.
\label{eq:head-common-ball}
\end{equation}

Since the gradient rows sum to zero, translating every update row by the same vector leaves the objective unchanged.
Using this freedom, we invoke a change of variable $\widehat {\Delta W}=\Delta W-\one\zeta^\top$, which expresses each row relative to the ball's center.
Each row $\widehat {\Delta w}_i$ then lies in $\mathcal B(0,\rho_{\mathrm{head}}/2)$.
Equation~\ref{eq:head-common-ball} therefore has the same minimum as
\begin{equation}
\min_{\widehat {\Delta W}} \inner{G}{\widehat {\Delta W}}_F
\qquad
\text{subject to}
\qquad
\widehat {\Delta w}_i\in\mathcal B(0,\rho_{\mathrm{head}}/2)
\quad\forall i.
\label{eq:head-origin-ball}
\end{equation}
Conversely, any $\widehat {\Delta W}$ satisfying this constraint gives a feasible common-ball update by choosing $\Delta W=\widehat {\Delta W}$ and $\zeta=0$.

A natural curiosity would be the suboptimality gap introduced by the tighter ball constraint.
Let $\Delta_\star$ and $\Delta_{\mathrm{ball}}$ denote the greatest first-order decreases under the diameter and common-ball constraints, respectively.
These are the negatives of the minimum objective values in Equations~\ref{eq:head-diameter-descent} and~\ref{eq:head-common-ball}.
For $G\neq0$ and $\rho_{\mathrm{head}}>0$, let $k=\operatorname{rank}(G)$.
Jung's theorem \citep{jung1901ueber} bounds the radius needed to enclose a set of rows with a given diameter.
Applying it to the two optimization problems bounds the relative suboptimality gap by
\begin{equation}
0\leq\frac{\Delta_\star-\Delta_{\mathrm{ball}}}{\Delta_\star}
\leq1-\sqrt{\frac{k+1}{2k}}.
\label{eq:head-rownorm-guarantee}
\end{equation}
The bound is sharp.
Appendix~\ref{app:head-rownorm} provides its proof and elaborates on the equivalence brought about by the change of variable.

To find an update with this guarantee, we solve Equation~\ref{eq:head-origin-ball} one row at a time.
Each row has its own constraint and contributes independently to the objective.
For a feasible row $\widehat {\Delta w}_i$, the Cauchy--Schwarz inequality and the radius constraint give
\[
\inner{g_i}{\widehat {\Delta w}_i}
\geq-\norm{g_i}_2\norm{\widehat {\Delta w}_i}_2
\geq-\frac{\rho_{\mathrm{head}}}{2}\norm{g_i}_2.
\]
For $g_i\neq0$, equality is attained by choosing $\widehat {\Delta w}_i$ opposite to $g_i$ with norm $\rho_{\mathrm{head}}/2$.
For a zero gradient row, we choose $\widehat {\Delta w}_i=0$.
Let $\mathsf{RN}$ normalize each nonzero row to unit Euclidean norm and leave zero rows unchanged.
Choosing ${\Delta W}=\widehat {\Delta W}$ and $\zeta=0$, we obtain one minimizing update for the common-ball problem,
\begin{equation}
{\Delta W}_{\mathrm{RN}}=-\frac{\rho_{\mathrm{head}}}{2}\mathsf{RN}(G).
\label{eq:head-rownorm}
\end{equation}
This update is feasible for the original diameter problem and attains the common-ball optimum.
Its first-order decrease is $\Delta_{\mathrm{ball}}$, so Equation~\ref{eq:head-rownorm-guarantee} applies directly to RowNorm.

A practical version subtracts the mean row from $\Delta W_{\mathrm{RN}}$ to obtain projected RowNorm.
This common translation preserves the objective, the row diameter, and the output probabilities.
Our Hilbert distance argument therefore independently recovers the direction derived by \citet{lau2026symmetrycompatibleprincipleoptimizerdesign} from the symmetries of the LM head.
Appendix~\ref{app:head-practical} gives an implementation with momentum and numerical stabilization.



\section{The attention module}
\label{sec:compositional-muon}

We extend our principle from the LM head to attention. Attention presents a more intricate setting: its softmax probabilities arise from bilinear scores that jointly depend on the query and key weights.
Bounding the Hilbert distance between attention distributions leads us to the QK constraint underlying Compositional Muon.

\paragraph{Attention distributions.}
Attention uses softmax to distribute probability across the positions visible to each query.
We study how this distribution changes when the query and key weights are updated, holding the input hidden states $h_1,\ldots,h_T\in\R^d$ fixed.
For a single head with query and key dimension $d_h$, the weights $W_Q,W_K\in\R^{d\times d_h}$ determine the attention scores through
\[
M=W_QW_K^\top\in\R^{d\times d},
\qquad
s_{ij}=\frac{h_i^\top M h_j}{\sqrt{d_h}}.
\]
We consider scaled dot-product attention without rotary position transforms or query--key normalization.
Under causal masking, let $p_{ij}$ denote the probability assigned by query $i$ to position $j\in\mathcal J_i=\{1,\ldots,i\}$.
We measure Hilbert distance on this visible set, where all probabilities are positive.

\paragraph{Hilbert distance bounds changes in score differences.}
Let $\Delta M$ denote the change in the QK product, so that the updated product is $M+\Delta M$.
At position $i$, write $p_i=(p_{ij})_{j\in\mathcal J_i}$ for the original attention distribution and $p_i^+$ for the distribution after this change.
Holding the hidden states fixed, the score $s_{ij}$ changes by
\[
\Delta s_{ij}
=\frac{h_i^\top(M+\Delta M)h_j-h_i^\top M h_j}{\sqrt{d_h}}
=\frac{h_i^\top\Delta M h_j}{\sqrt{d_h}}.
\]
Within row $i$, the score difference between positions $j$ and $k$ therefore changes by $h_i^\top\Delta M(h_j-h_k)/\sqrt{d_h}$.
Applying Equation~\ref{eq:hilbert-softmax} within each attention row gives
\begin{equation}
\mathcal H(p_i,p_i^+)
=\max_{j,k\in\mathcal J_i}\abs{\Delta s_{ij}-\Delta s_{ik}}
=\frac{1}{\sqrt{d_h}}\max_{j,k\in\mathcal J_i}\abs{h_i^\top\Delta M(h_j-h_k)}.
\label{eq:attention-hilbert-comparisons}
\end{equation}
Intuitively, the probability constraint, specific to the attention mechanism, asks how much an update can change a query's preference between two keys.

\paragraph{The induced spectral norm constraint.}
Assume that every hidden state satisfies $\norm{h_\ell}_2\leq R$, with $R=\sqrt d$ corresponding to the RMS unit ball.
The hidden state $h_i$ has norm at most $R$, while the difference $h_j-h_k$ has norm at most $2R$.
Consequently,
\(
\abs{h_i^\top\Delta M(h_j-h_k)}
\leq\norm{h_i}_2\norm{\Delta M}_{\mathrm{op}}\norm{h_j-h_k}_2
\leq2R^2\norm{\Delta M}_{\mathrm{op}}.
\)

For causal sequences with $T\geq3$, the worst-case Hilbert distance over this input set is
\begin{equation}
\sup_{\substack{\norm{h_\ell}_2\leq R\\1\leq\ell\leq T}}
\max_{1\leq i\leq T}\mathcal H(p_i,p_i^+)
=\frac{2R^2}{\sqrt{d_h}}\norm{\Delta M}_{\mathrm{op}}.
\label{eq:attention-hilbert-spectral}
\end{equation}
Requiring a Hilbert distance of at most $\varepsilon$ for every attention row throughout this input set is therefore equivalent to
\begin{equation}
\norm{\Delta M}_{\mathrm{op}}\leq\rho_{\mathrm{QK}},
\qquad
\rho_{\mathrm{QK}}=\frac{\varepsilon\sqrt{d_h}}{2R^2}.
\label{eq:attention-product-constraint}
\end{equation}

\emph{Why does the same Hilbert distance principle give a row diameter for the LM head and a spectral norm for attention?}
In the LM head, each comparison involves a single hidden state, whereas attention involves a query input and a difference between key inputs.
Under the stated norm bounds, these two vectors can be aligned independently to attain the spectral norm of $\Delta M$.
The bound is attained when $h_i$ is $R$ times a unit leading left singular vector of $\Delta M$, and $h_j-h_k$ is $2R$ times the corresponding unit right singular vector.

\paragraph{Steepest descent over query and key updates.}
Let $\Delta W_Q$ and $\Delta W_K$ denote updates to the query and key weights $W_Q$ and $W_K$, respectively.
Expanding the updated product gives
\[
\Delta M
=
(W_Q+\Delta W_Q)(W_K+\Delta W_K)^\top-W_QW_K^\top
=
\underbrace{\Delta W_QW_K^\top+W_Q\Delta W_K^\top}_{\Delta M_{\mathrm{lin}}}
+\Delta W_Q\Delta W_K^\top.
\]
The first two terms give the first-order change $\Delta M_{\mathrm{lin}}(\Delta W_Q,\Delta W_K)$, while $\Delta W_Q\Delta W_K^\top$ is second order in the updates and is therefore omitted.
Applying Equation~\ref{eq:attention-product-constraint} to this first-order change gives
\[
\norm{\Delta W_QW_K^\top+W_Q\Delta W_K^\top}_{\mathrm{op}}
\leq\rho_{\mathrm{QK}}.
\]
We assume that these weights affect the loss only through the attention probabilities.
Write $G_Q=\nabla_{W_Q}\mathcal L$ and $G_K=\nabla_{W_K}\mathcal L$ for their gradients.
Constructing our steepest descent problem under this Hilbert distance bound, we obtain
\begin{equation}
\begin{aligned}
\min_{\Delta W_Q,\Delta W_K}\quad
&\inner{G_Q}{\Delta W_Q}_F+\inner{G_K}{\Delta W_K}_F\\
\text{subject to}\quad
&\norm{\Delta W_QW_K^\top+W_Q\Delta W_K^\top}_{\mathrm{op}}
\leq\rho_{\mathrm{QK}}.
\end{aligned}
\label{eq:attention-compositional-descent}
\end{equation}
Equation~\ref{eq:attention-compositional-descent} is precisely the first-order QK steepest descent problem posed by \citet[Section~1.2]{keigwin2026compositional}.
Their derivation controls changes in the composed bilinear map through the spectral norm of its first-order perturbation.
Our starting point is a bound on the Hilbert distance between attention distributions.
Translating this probability bound through softmax and the bilinear scores recovers the same constraint on the first-order QK product change.
Our Hilbert distance derivation therefore provides a natural prelude to Compositional Muon, motivating its QK constraint through the stability of the attention distributions.

The LM head and attention affect optimization through both forward probability distributions and the gradients propagated to the hidden states.
Our main analysis bounds changes in the former and derives the corresponding steepest descent problems.
Holding the inputs and incoming gradients fixed, Appendix~\ref{app:perturbation-bounds} bounds the resulting perturbations of the latter.


\section{Discussion}
\label{sec:discussion}
\textit{Can Muon be adapted to softmax simply by removing the common shifts that softmax ignores?}
Because softmax is invariant to common shifts, one might try adapting Muon by centering the update rows before measuring their spectral norm.
We show that the resulting steepest descent problem has the same minimum and that the Muon update is already row centered.
Translation symmetry alone therefore does not alter Muon, motivating the stronger relative control provided by Hilbert distance.

\paragraph{Centering leaves Muon unchanged.}

Consider a linear module with $n$ outputs and a parameter matrix whose influence on the loss is entirely through the softmax output distribution.
Let $P=I_n-\one\one^\top/n$ denote the orthogonal projection that subtracts the mean row.
Since gradient rows sum to zero, $PG=G$.
Since $P$ is an orthogonal projection, $\norm{P\Delta W}_{\mathrm{op}}\leq\norm{P}_{\mathrm{op}}\norm{\Delta W}_{\mathrm{op}}\leq\norm{\Delta W}_{\mathrm{op}}$.
Thus, every update satisfying $\norm{\Delta W}_{\mathrm{op}}\leq\rho$ also satisfies $\norm{P\Delta W}_{\mathrm{op}}\leq\rho$.
Conversely, take any update $\Delta W$ satisfying $\norm{P\Delta W}_{\mathrm{op}}\leq\rho$.
The centered update $P\Delta W$ has spectral norm at most $\rho$.
It also preserves the objective, since $\inner{G}{P\Delta W}_F=\inner{PG}{\Delta W}_F=\inner{G}{\Delta W}_F$.
The additional freedom therefore leaves the minimum unchanged,
\[
\min_{\norm{P\Delta W}_{\mathrm{op}}\leq\rho}\inner{G}{\Delta W}_F
=
\min_{\norm{\Delta W}_{\mathrm{op}}\leq\rho}\inner{G}{\Delta W}_F.
\]
The Muon update is already centered.
Indeed, if $G=A\Sigma B^\top$ is the compact SVD of a nonzero gradient, then every left singular vector satisfies
$\one^\top a_j=(\one^\top G b_j)/\sigma_j=0$, giving
\[
\Delta W_{\mathrm{Muon}}=-\rho AB^\top,
\qquad
P\Delta W_{\mathrm{Muon}}=-\rho(PA)B^\top=\Delta W_{\mathrm{Muon}}.
\]
Accounting for translation symmetry alone therefore recovers the standard Muon solution.
Our Hilbert distance approach goes further by bounding the largest change in the difference between any two preactivation coordinates.
For a centered preactivation perturbation $\Delta z\in\R^n$, with $p=\softmax(z)$ and $q=\softmax(z+\Delta z)$, we have
$2\norm{\Delta z}_{\mathrm{RMS}}
\leq
\mathcal H(p,q)
\leq
\sqrt{2n}\,\norm{\Delta z}_{\mathrm{RMS}}$.
Since the upper bound is sharp, a fixed bound on centered RMS permits Hilbert distance to grow with the output width.
Appendix~\ref{app:rms-hilbert-comparison} proves these bounds and discusses their sharpness.

\paragraph{Limitations.}
The main objective of this work is to provide a unifying mathematical foundation for softmax modules.
Through Hilbert distance, we independently recover RowNorm for the LM head and the first-order QK constraint underlying Compositional Muon.
These mathematical connections naturally raise the question of what this perspective offers in practice.
We leave that question to future experimental work and make no empirical claim of improved training speed or final model quality.
The work of \citet{lau2026symmetrycompatibleprincipleoptimizerdesign} on RowNorm and that of \citet{keigwin2026compositional} on Compositional Muon report improvements in language-model pretraining and include ablations of their optimizer choices.
These experiments provide existing empirical evidence for the practical relevance of the methods connected by our framework.

A second limitation arises from the relative nature of Hilbert distance.
Because it controls pairwise odds, it can be sensitive to large relative changes in low probability outcomes even when little probability mass moves.
For example, suppose an update raises the probability of one outcome from $10^{-9}$ to $10^{-6}$ while leaving the rest of the distribution nearly unchanged.
The probability of this outcome increases by a factor of $1000$, producing a large Hilbert distance despite the small movement of probability mass,
\(
\mathrm{TV}(p,q)=9.99\times10^{-7},
\mathcal H(p,q)\approx6.91.
\)
A small Hilbert distance constraint can therefore exclude updates that move very little probability mass.
Future work could investigate metrics that preserve the useful guarantees developed here while being less sensitive to low probability outcomes.

\section{Related work}
\label{sec:related-work}

\paragraph{From Adam to module specific optimizers.}
Adam has long served as the default optimizer across deep learning \citep{kingma2014adam}.
Muon has recently challenged this convention \citep{jordan2024muon}.
For a linear layer, Muon controls the worst-case change in its output as measured by RMS, resulting in a bound on the spectral norm of the weight update.
The resulting steepest descent problem forms the mathematical foundation of the Muon update \citep{bernstein2025deriving}.
The Modula project generalizes this perspective and advocates constructing optimizers whose geometry reflects the function of each module
\citep{modula-docs,large2024scalableoptimizationmodularnorm}.
Modular duality and Scion provide systematic ways to construct updates from such geometries
\citep{bernstein2024modular,pethick2025trainingdeeplearningmodels}.
We extend this principle to modules involving softmax.

\paragraph{Geometry on probability distributions.}
A related line of work connects parameter updates to the probability distributions they induce.
Natural gradient equips a family of probability distributions with the Fisher geometry and chooses updates according to the resulting notion of steepest descent
\citep{amari1998natural,martens2020new}.
TRPO makes this connection explicit by constraining the expected KL divergence between the action distributions induced by the original and updated policies
\citep{schulman2015trust}.
Hilbert's projective metric offers a different geometry for positive vectors.
Beginning with the contraction theory of positive operators, this geometry has subsequently appeared in matrix scaling, clustering, nonlinear embeddings, and the study of probability simplices
\citep{birkhoff1957extensions,de1993hilbert,cuturi2013sinkhorn,Nielsen_2018,nielsen2023non}.
More recently, \citet{cohen2024hyperboliccontractivityhilbertmetric} established a sharp relationship between Hilbert distance and total variation.
We draw on this literature to study the parameter geometry induced by Hilbert distance on the probability distributions produced by softmax.

\paragraph{Optimizers for the LM head and attention.}
The LM head and attention have each motivated specialized optimization methods.
For the LM head, Scion \citep{pethick2025trainingdeeplearningmodels} includes RowNorm and SCALE \citep{glentis2026memoryefficientllmpretrainingminimalist} proposes a momentum-based variant; \citet{lau2026symmetrycompatibleprincipleoptimizerdesign} derive projected RowNorm from its permutation and shared-shift symmetries.
Attention has motivated a different collection of interventions.
Query and key normalization controls the scale of attention scores, while spectral reparameterization addresses attention entropy collapse
\citep{henry2020query,zhai2023stabilizing}.
MuonClip monitors the maximum attention logits and conditionally rescales the query and key projection weights after a Muon update
\citep{kimiteam2026kimik2openagentic}.
Compositional Muon goes further by constraining the first-order change in the product of the query and key matrices
\citep{keigwin2026compositional}.
Although row normalization for the LM head and product control for attention appear to be independent design choices, we show that they share common ground in bounding changes in the probability distributions produced by softmax.

\section{Conclusion}

We develop a unified framework for optimizing composed softmax modules by bounding changes in their output probabilities and deriving the corresponding parameter constraints.
This addresses a limitation of Muon's RMS control, which can conceal substantial probability redistribution when applied to the intermediate linear output.
We propose using Hilbert distance to constrain these changes in the output distribution directly.
Hilbert distance respects the common-shift invariance of softmax and controls both the redistribution of probability mass and changes in cross-entropy loss.
Its representation as the variation seminorm of the preactivation perturbation lets us translate a common constraint on the output distribution into a parameter geometry for each module.
For the LM head, this gives a row diameter constraint.
A tighter ball approximation leads to RowNorm, with centering giving its projected form.
For attention, the same principle recovers the first-order QK steepest descent problem underlying Compositional Muon.
A single probability constraint thus connects the separately developed RowNorm update and Compositional Muon's QK optimization problem.
We offer this framework as a contribution to the broader Modula perspective, demonstrating how the output geometry of composed softmax modules can guide optimizer design.
We hope these connections spark new ideas for optimizers and architectures built around composed modules.
Hilbert distance provides a starting point for softmax, and we encourage researchers to ask which output geometries could play that role for other compositions.

\newpage

\subsection*{AI use statement}




The idea of bounding changes in softmax output probabilities originated with the authors. The choice of Hilbert distance was found by studying many existing distances in the literature. AI-assisted mathematical exploration also informed the selection of Hilbert distance. AI tools were extensively used to survey literature, identify related work, check mathematical arguments, and verify correctness. In particular, the authors initially obtained a weaker approximation bound for RowNorm relative to the diameter-constrained optimum. AI tools subsequently identified Jung's theorem and its applicability in tightening this bound.

The initial manuscript was written by the authors. AI tools were subsequently used to improve wording, technical exposition, and presentation, and to provide critical feedback on the manuscript. The authors reviewed all AI assisted editorial changes before incorporating them. We, the authors, take responsibility for the final content of the paper, including its mathematical claims, proofs, and references.

\subsection*{Ethics statement}



This work develops a unified framework for optimizing modules which involve softmax. The research presented here does not involve human participants, personal data, or the collection or release of datasets. The framework is general purpose, and all ethical considerations in downstream applications depend on the training data, models, and deployment contexts in which it is used.

\subsection*{Reproducibility statement}



The assumptions underlying our theoretical results are stated alongside the derivations in the main text. Appendix~\ref{app:proofs} provides the sharpness argument for the total variation bound, the exact dual formulation of the diameter-constrained problem, and the proof of the sharp RowNorm approximation guarantee. Appendix~\ref{app:perturbation-bounds} derives the forward and backward perturbation bounds for the LM head and attention.
Appendix~\ref{app:head-practical} specifies the practical RowNorm update, including centering, momentum, and numerical stabilization. To reproduce Figure~\ref{fig:rms-softmax-distributions}, Appendix~\ref{app:figure-construction} gives deterministic formulas for the initial logits and all three updates, together with their RMS norms, Hilbert distances, and total variation values.

\subsubsection*{Author Contributions}
If you'd like to, you may include  a section for author contributions as is done
in many journals. This is optional and at the discretion of the authors.

\subsubsection*{Acknowledgments}
Use unnumbered third level headings for the acknowledgments. All
acknowledgments, including those to funding agencies, go at the end of the paper.

\bibliography{iclr2027_conference}
\bibliographystyle{iclr2027_conference}

\appendix

\section{Proofs and further connections}
\label{app:proofs}

\subsection{Comparing centered RMS and Hilbert distance}
\label{app:rms-hilbert-comparison}

This subsection proves the comparison used in Section~\ref{sec:discussion}. For centered preactivation perturbations, Hilbert distance bounds RMS independently of the output width, whereas the reverse bound necessarily grows with the width.

Let $z\in\R^n$ be a preactivation vector and let $\Delta z\in\R^n$ satisfy $\one^\top\Delta z=0$. Define $p=\softmax(z)$, $q=\softmax(z+\Delta z)$, and $\norm{\Delta z}_{\mathrm{RMS}}=\norm{\Delta z}_2/\sqrt n$. We prove
\begin{equation}
2\norm{\Delta z}_{\mathrm{RMS}}
\leq
\mathcal H(p,q)
\leq
\sqrt{2n}\,\norm{\Delta z}_{\mathrm{RMS}}.
\label{eq:rms-hilbert-comparison}
\end{equation}

The softmax identity gives $\mathcal H(p,q)=\max_{i,j}\abs{\Delta z_i-\Delta z_j}$.
Applying the Cauchy--Schwarz inequality gives
\[
\abs{\Delta z_i-\Delta z_j}
\leq\sqrt2\sqrt{\Delta z_i^2+\Delta z_j^2}
\leq\sqrt{2\sum_{k=1}^n\Delta z_k^2}
=\sqrt{2n}\norm{\Delta z}_{\mathrm{RMS}},
\qquad i\neq j.
\]
Taking the maximum over $i$ and $j$ proves the upper bound. Equality is attained by $\Delta z=(a,-a,0,\ldots,0)$ for any $a>0$, so the factor $\sqrt{2n}$ is sharp.

For the lower bound, let $m=\min_i\Delta z_i$, $M=\max_i\Delta z_i$, and $c=(m+M)/2$. Since $\sum_i\Delta z_i=0$,
\[
\norm{\Delta z}_{\mathrm{RMS}}^2
\leq\norm{\Delta z}_{\mathrm{RMS}}^2+c^2
=\frac1n\sum_i(\Delta z_i-c)^2
\leq\max_i(\Delta z_i-c)^2
=\left(\frac{M-m}{2}\right)^2
=\frac{\mathcal H(p,q)^2}{4}.
\]
Dropping the nonnegative term $c^2$ proves the lower bound. When $n$ is even, equality is attained by taking half the coordinates equal to $a$ and half equal to $-a$.

Equation~\ref{eq:rms-hilbert-comparison} reveals an asymmetry between the two measures. Hilbert distance bounds centered RMS with a constant independent of $n$, but a fixed RMS bound permits Hilbert distance to grow as $\sqrt n$. A similar dependence appears in the standard comparison between the spectral and Frobenius norms: for a matrix $A$ of rank $r$,
\[
\norm{A}_{\mathrm{op}}
\leq
\norm{A}_F
\leq
\sqrt r\,\norm{A}_{\mathrm{op}}.
\]
In both comparisons, an aggregate Euclidean measure can conceal an extremal change whose magnitude grows with dimension or rank. Appendix~\ref{app:softmax-tv} shows that, for softmax, this separation can persist at the level of the output distribution: centered RMS can vanish while total variation approaches $1/2$.

\subsection{Probability redistribution under softmax}
\label{app:softmax-tv}

This subsection establishes two consequences of the Hilbert geometry developed in the main text. First, we show that the total variation bound is sharp for every fixed preactivation perturbation. We then construct centered perturbations whose RMS tends to zero while their probability redistribution remains bounded away from zero as the output width grows.

\paragraph{The sharp total variation bound.}
Recall that the variation seminorm is $\gamma=\varnorm{\Delta z}=\max_\ell\Delta z_\ell-\min_\ell\Delta z_\ell$.
For $p=\softmax(z)$ and $q=\softmax(z+\Delta z)$, Equation~\ref{eq:hilbert-softmax} and the sharp bound of \citet[Theorem~5.1]{cohen2024hyperboliccontractivityhilbertmetric} give $\mathrm{TV}(p,q)\leq\tanh(\gamma/4)$.
To approach equality when $\gamma>0$, choose $i,j$ with $\Delta z_i-\Delta z_j=\gamma$ and set $z_i=-\gamma/2$, $z_j=0$.
The difference between these preactivation coordinates therefore changes from $-\gamma/2$ to $\gamma/2$.
Let all other preactivations tend to $-\infty$, concentrating both distributions on $i,j$.
The probability gained by coordinate $i$ then gives
\[
\mathrm{TV}(p,q)\longrightarrow
\frac{e^{\gamma/2}}{1+e^{\gamma/2}}
-\frac{1}{1+e^{\gamma/2}}
=\tanh\!\left(\frac{\gamma}{4}\right).
\]
This approaches the upper bound through strictly positive softmax distributions, proving the claimed supremum.
For $\gamma=0$, the update is a common shift and both sides are zero.

\paragraph{Vanishing RMS with nonvanishing probability redistribution.}
We now show that centering alone does not make RMS a uniform measure of probability change.
Take $n\geq3$, initial preactivation vector $z=0$, and $a_n=\log(n-1)$.
The initial probabilities and centered update are
\[
p=\frac{\one}{n},
\qquad
\Delta z=a_n\left(1-\frac1n,-\frac1n,\ldots,-\frac1n\right)^\top,
\qquad
\one^\top\Delta z=0.
\]
By softmax's common-shift invariance, $q=\softmax(\Delta z)$ satisfies
\[
q_1=\frac{e^{a_n}}{e^{a_n}+n-1}=\frac12,
\qquad
q_i=\frac{1}{e^{a_n}+n-1}=\frac{1}{2(n-1)}
\quad(i>1).
\]
Yet, as $n\to\infty$,
\[
\norm{\Delta z}_{\mathrm{RMS}}=\frac{\sqrt{n-1}}{n}\log(n-1)\longrightarrow0,
\qquad
\mathrm{TV}(p,q)=\frac12-\frac1n\longrightarrow\frac12.
\]
Even after centering, vanishing RMS does not ensure vanishing probability redistribution uniformly across widths.
Meanwhile, $\mathcal H(p,q)=\log(n-1)$ captures the growing relative change.

\subsection{The exact dual of the diameter constraint}
\label{app:head-duality}

This subsection derives the exact dual of the row diameter problem in Equation~\ref{eq:head-diameter-descent}. The dual represents each gradient row through antisymmetric vectors assigned to pairs of vocabulary rows and minimizes their total Euclidean norm. We first establish the balance constraints and the dual identity, then relate the resulting problem to vector transport.

Let $G\in\R^{V\times d}$ have rows $g_i^\top$, where $V\geq2$ and $\sum_i g_i=0$. For $\rho_{\mathrm{head}}>0$, consider
\begin{equation}
\min_{\Delta W}
\inner{G}{\Delta W}_F
\qquad
\text{subject to}
\qquad
\norm{\Delta w_i-\Delta w_j}_2
\leq
\rho_{\mathrm{head}}
\quad
\forall i<j.
\label{eq:head-diameter-primal-app}
\end{equation}

\paragraph{Pairwise dual variables.}
For every pair $i<j$, introduce a vector $\varphi_{ij}\in\R^d$. Extend this collection by setting $\varphi_{ji}=-\varphi_{ij}$ and $\varphi_{ii}=0$. The dual feasibility conditions are
\begin{equation}
\sum_{j=1}^V\varphi_{ij}=g_i,
\qquad
i=1,\ldots,V.
\label{eq:head-flow-balance}
\end{equation}
Antisymmetry implies
\[
\sum_i\sum_j\varphi_{ij}
=
\sum_{i<j}(\varphi_{ij}+\varphi_{ji})
=
0.
\]
Hence, the condition $\sum_i g_i=0$ is necessary for Equation~\ref{eq:head-flow-balance}. It is also sufficient. For example, setting $\varphi_{iV}=g_i$ for $i<V$ and every other independently indexed variable to zero satisfies all balance equations because $g_V=-\sum_{i<V}g_i$.

\paragraph{The exact dual identity.}
The primal problem in Equation~\ref{eq:head-diameter-primal-app} satisfies
\begin{equation}
\min_{D(\Delta W)\leq\rho_{\mathrm{head}}}
\inner{G}{\Delta W}_F
=
-\rho_{\mathrm{head}}
\min_{\substack{\Phi\\
\sum_j\varphi_{ij}=g_i\;\forall i}}
\sum_{i<j}\norm{\varphi_{ij}}_2.
\label{eq:head-diameter-dual}
\end{equation}
Both minima are attained. Consequently, the greatest first order decrease under the diameter constraint is
\[
\Delta_\star
=
\rho_{\mathrm{head}}
\min_{\substack{\Phi\\
\sum_j\varphi_{ij}=g_i\;\forall i}}
\sum_{i<j}\norm{\varphi_{ij}}_2.
\]

\begin{proof}

We first establish attainment of the primal minimum. Given any feasible $\Delta W$, subtract $\Delta w_V$ from every update row. This translation preserves all pairwise differences and changes the objective by
\[
-\inner{\sum_i g_i}{\Delta w_V}
=
0.
\]
We may therefore restrict attention to representatives satisfying $\Delta w_V=0$. Feasibility then gives $\norm{\Delta w_i}_2\leq\rho_{\mathrm{head}}$ for every $i<V$. The resulting feasible set is nonempty and compact, so the continuous objective attains a finite minimum.


To derive the dual, introduce an auxiliary variable $\xi_{ij}$ ffor each pair $i<j$ and impose
\[
\xi_{ij}=\Delta w_i-\Delta w_j,
\qquad
\norm{\xi_{ij}}_2\leq\rho_{\mathrm{head}},
\qquad i<j.
\]
Associate the multiplier $\varphi_{ij}$ with the equality constraint $\xi_{ij}-\Delta w_i+\Delta w_j=0$. The Lagrangian is
\[
\begin{aligned}
\Lambda(\Delta W,\xi,\Phi)
&=
\inner{G}{\Delta W}_F
+
\sum_{i<j}
\inner{\varphi_{ij}}
{\xi_{ij}-\Delta w_i+\Delta w_j}\\
&=
\sum_{i=1}^V
\inner{g_i-\sum_{j=1}^V\varphi_{ij}}
{\Delta w_i}
+
\sum_{i<j}
\inner{\varphi_{ij}}{\xi_{ij}}.
\end{aligned}
\]
The rows of $\Delta W$ are unrestricted in the inner minimization. Its infimum is therefore finite only if
\(
\sum_j\varphi_{ij}=g_i
\)
for all $i$.
Under these balance equations, the terms containing $\Delta W$ vanish. Minimizing over the Euclidean balls gives
\[
\inf_{\substack{\xi\\
\norm{\xi_{ij}}_2\leq\rho_{\mathrm{head}}\;\forall i<j}}
\sum_{i<j}\inner{\varphi_{ij}}{\xi_{ij}}
=
-\rho_{\mathrm{head}}
\sum_{i<j}\norm{\varphi_{ij}}_2.
\]
For $\varphi_{ij}\neq0$, the infimum for that pair is attained at
\[
\xi_{ij}
=
-\rho_{\mathrm{head}}
\frac{\varphi_{ij}}{\norm{\varphi_{ij}}_2};
\]
when $\varphi_{ij}=0$, any feasible $\xi_{ij}$ is optimal.
The Lagrange dual therefore maximizes
\[
-\rho_{\mathrm{head}}
\sum_{i<j}\norm{\varphi_{ij}}_2
\]
subject to Equation~\ref{eq:head-flow-balance}. Since $-\rho_{\mathrm{head}}<0$, this is equivalent to the minimization on the right hand side of Equation~\ref{eq:head-diameter-dual}.

The reformulated primal problem is convex. Moreover, $\Delta W=0$ and $\xi_{ij}=0$ satisfy every equality constraint and lie strictly inside every Euclidean ball. Slater's condition therefore gives strong duality and attainment of the dual optimum \citep[Section~5.2.3]{vandenberghe2004convex}. This proves Equation~\ref{eq:head-diameter-dual}.
\end{proof}



\paragraph{Connection to vector transport.}

Let $\mathcal X=\{1,\ldots,V\}$ carry the discrete metric $d(i,j)=1$ for $i\neq j$, and define the vector measure
\[
\mu
=
\sum_{i=1}^V g_i\delta_i.
\]
The condition $\sum_i g_i=0$ gives $\mu(\mathcal X)=0$. Equation~\ref{eq:head-flow-balance} represents this measure through an antisymmetric vector flow on the complete graph, with transport cost
\[
\sum_{i<j}\norm{\varphi_{ij}}_2.
\]
The primal variables provide the corresponding Lipschitz functions. In particular, define
\[
u_i
=
-\frac{\Delta w_i}{\rho_{\mathrm{head}}}.
\]
The diameter constraint implies
\[
\norm{u_i-u_j}_2
\leq
1
=
d(i,j)
\qquad
\forall i\neq j,
\]
so $u:\mathcal X\to\R^d$ is $1$ Lipschitz. Equation~\ref{eq:head-diameter-dual} is therefore the finite discrete form of vector transport duality \citep[Theorem~2]{ciosmak2021optimal}.

This dual formulation characterizes the optimal value without requiring the primal minimizer to be unique. It contains one vector in $\R^d$ for each of the $\binom V2$ unordered pairs of vocabulary rows. The resulting quadratic growth in the number of dual variables motivates the common ball approximation studied next.



\subsection{The RowNorm approximation}
\label{app:head-rownorm}

The main text obtains RowNorm by requiring all update rows to lie in a common ball and then moving that ball to the origin.
We first justify why this change of variable preserves the minimum.
We then bound the suboptimality introduced by replacing the diameter constraint with the ball constraint.

\subsubsection{Centering the enclosing ball preserves the optimum}
\label{app:head-ball-center}

Fix a gradient $G\in\R^{V\times d}$ with rows $g_i^\top$ satisfying $\sum_i g_i=0$, and let $\rho_{\mathrm{head}}>0$.
Recall that $\mathcal B(\zeta,\rho_{\mathrm{head}}/2)$ denotes the closed Euclidean ball with center $\zeta$ and radius $\rho_{\mathrm{head}}/2$.
In Equation~\ref{eq:head-common-ball}, we choose both the update $\Delta W$ and the center $\zeta$,
\[
\min_{\Delta W,\zeta}\inner{G}{\Delta W}_F
\qquad
\text{subject to}
\qquad
\Delta w_i\in\mathcal B(\zeta,\rho_{\mathrm{head}}/2)
\quad\forall i.
\]
Any center that encloses all the update rows is allowed.
In particular, the center is not required to equal the mean row, and the mean row remains unrestricted.

Since the gradient rows sum to zero, translating every update row by the same vector leaves the objective unchanged.
Using this freedom, we invoke a change of variable
\[
\widehat{\Delta W}
=
\Delta W-\one\zeta^\top.
\]
For any feasible pair $(\Delta W,\zeta)$, the translated rows satisfy
\[
\norm{\widehat{\Delta w}_i}_2
=
\norm{\Delta w_i-\zeta}_2
\leq
\frac{\rho_{\mathrm{head}}}{2}
\qquad\forall i.
\]
Thus, every row now lies in a ball centered at the origin.
The objective is preserved because
\[
\inner{G}{\Delta W}_F
=
\inner{G}{\widehat{\Delta W}}_F
+
\underbrace{
\Bigl\langle\sum_i g_i,\zeta\Bigr\rangle
}_{=0}
=
\inner{G}{\widehat{\Delta W}}_F.
\]
After this change of variable, neither the row constraints nor the objective depends on $\zeta$.
Equation~\ref{eq:head-common-ball} therefore has the same minimum as
\[
\min_{\widehat{\Delta W}}
\inner{G}{\widehat{\Delta W}}_F
\qquad
\text{subject to}
\qquad
\widehat{\Delta w}_i\in
\mathcal B(0,\rho_{\mathrm{head}}/2)
\quad\forall i,
\]
which is Equation~\ref{eq:head-origin-ball}.
To complete the equivalence, we also need to return from this problem to the original one.
Given any feasible $\widehat{\Delta W}$, choose any $\zeta\in\R^d$ and set
\[
\Delta W
=
\widehat{\Delta W}+\one\zeta^\top.
\]
The resulting rows lie in $\mathcal B(\zeta,\rho_{\mathrm{head}}/2)$, and the same objective identity shows that their value is unchanged.
Choosing $\zeta=0$ simply gives $\Delta W=\widehat{\Delta W}$.
The two problems therefore attain exactly the same objective values.

The feasible set for the problem centered at the origin is a finite product of closed, bounded balls.
It is nonempty and compact, so the continuous linear objective attains its minimum.
The reverse construction then provides a minimizing pair for the original problem as well.
Consequently, every minimizer of the problem centered at the origin gives a minimizer of the common-ball problem, and every minimizing pair of the common ball problem translates back to one.

This change of variable centers the enclosing ball.
The rows inside it can still have a nonzero mean.
The RowNorm update in Equation~\ref{eq:head-rownorm} is therefore already one minimizing choice for the common ball problem, with $\zeta=0$.
Subtracting its mean row is optional and preserves the objective by the same translation argument.

Moving the ball to the origin introduces no additional suboptimality.
The gap considered next comes entirely from requiring the update rows to fit inside a ball of radius $\rho_{\mathrm{head}}/2$ instead of imposing only the original diameter constraint.

\subsubsection{Approximation guarantee and sharpness}
\label{app:head-rownorm-gap}

Moving the ball to the origin preserves the minimum.
We now bound the first-order decrease lost by replacing the diameter constraint with the common-ball constraint.
Assume $G\neq0$, $\one^\top G=0$, and $\rho_{\mathrm{head}}>0$.
Let $\Delta_\star$ and $\Delta_{\mathrm{ball}}$ denote the greatest first-order decreases under these two constraints, respectively.
Since RowNorm attains the common-ball optimum,
\[
\Delta_{\mathrm{ball}}=\frac{\rho_{\mathrm{head}}}{2}\sum_i\norm{g_i}_2>0.
\]

\paragraph{Bounding the gap.}
Jung's theorem bounds the radius needed to enclose a set using its diameter and dimension.
Here we can work in the space spanned by the gradient rows, $\mathcal S=\operatorname{span}\{g_1,\ldots,g_V\}$.
By definition of matrix rank, the dimension of this row space is $k=\operatorname{rank}(G)$.

To see why this space suffices, consider the part of an update row perpendicular to $\mathcal S$.
It has zero inner product with every gradient row and therefore contributes nothing to the objective.
Removing these parts from all rows is an orthogonal projection, which cannot increase pairwise distances.
Every feasible update can thus be replaced by one whose rows lie in $\mathcal S$, with the same objective value and no larger diameter.

Jung's theorem \citep{jung1901ueber} guarantees that any such rows with diameter at most $\rho_{\mathrm{head}}$ lie in a ball of radius at most
\[
r_k=\rho_{\mathrm{head}}\sqrt{\frac{k}{2(k+1)}}.
\]
Let $\zeta$ be the center of such a ball, so that $\norm{\Delta w_i-\zeta}_2\leq r_k$ for every row.
We only need to know that this center exists.
As in the preceding subsection, $\sum_i g_i=0$ allows us to subtract $\zeta$ without changing the objective.
Cauchy--Schwarz then gives
\[
-\inner{G}{\Delta W}_F=-\sum_i\inner{g_i}{\Delta w_i-\zeta}\leq r_k\sum_i\norm{g_i}_2.
\]
This bounds $\Delta_\star$ because every feasible update can be brought into $\mathcal S$ without changing its objective value.
Every common-ball update also satisfies the diameter constraint, so
\[
0<\Delta_{\mathrm{ball}}\leq\Delta_\star\leq\sqrt{\frac{2k}{k+1}}\,\Delta_{\mathrm{ball}}.
\]
Rearranging proves the relative suboptimality bound in Equation~\ref{eq:head-rownorm-guarantee}.

\paragraph{Sharpness.}
A regular simplex gives a gradient for which equality holds.
Choose $g_1,\ldots,g_{k+1}$ as the unit vertices of a regular $k$-dimensional simplex centered at the origin in $\R^d$.
For a vocabulary with $V>k+1$, set $g_i=0$ for every $i>k+1$.
The simplex rows satisfy
\[
\sum_{i=1}^{k+1}g_i=0,
\qquad
\norm{g_i}_2=1,
\qquad
\inner{g_i}{g_j}=-\frac1k
\quad(i\neq j).
\]
The full gradient therefore satisfies $\one^\top G=0$ and $\operatorname{rank}(G)=k$.
Consider the update $\Delta W=-r_kG$.
For every $i>k+1$, this sets $\Delta w_i=-r_kg_i=0$.
Distinct simplex update rows have distance
\[
\norm{\Delta w_i-\Delta w_j}_2=r_k\sqrt{\frac{2(k+1)}{k}}=\rho_{\mathrm{head}}.
\]
Each simplex update row is also at distance $r_k\leq\rho_{\mathrm{head}}$ from every zero update row.
The entire update is therefore feasible.
It achieves a decrease of $r_k(k+1)$, matching the upper bound on $\Delta_\star$.
Since $\Delta_{\mathrm{ball}}=\rho_{\mathrm{head}}(k+1)/2$, the relative gap is
\[
\frac{\Delta_\star-\Delta_{\mathrm{ball}}}{\Delta_\star}=1-\frac{\rho_{\mathrm{head}}}{2r_k}=1-\sqrt{\frac{k+1}{2k}}.
\]
This proves sharpness for every permissible rank $1\leq k\leq\min(d,V-1)$.
RowNorm therefore attains the optimal first-order decrease when $k=1$ and retains at least $1/\sqrt2$ of it for every rank.
The next subsection describes the practical update with momentum and numerical stabilization.

\subsection{Practical RowNorm}
\label{app:head-practical}
In training, we apply row normalization to a first-moment estimate of the gradient.
Let $P=I_V-\one\one^\top/V$ denote the projection that subtracts the mean row of a matrix.
Starting from $B_{-1}=0$, for $t=0,1,\ldots$ let
\[
B_t=\beta B_{t-1}+(1-\beta)G_t,
\qquad
\widehat B_t=\frac{B_t}{1-\beta^{t+1}},
\qquad
\widetilde B_t=P\widehat B_t,
\]
where $0\leq\beta<1$.
For numerical stability, define $\mathsf{RN}_{\delta_{\mathrm{num}}}$ by dividing row $\widetilde b_{t,i}^\top$ by $\norm{\widetilde b_{t,i}}_2+\delta_{\mathrm{num}}$, with $\delta_{\mathrm{num}}>0$.
With learning rate $\eta_t\geq0$, the update is
\begin{equation}
\Delta W_t=-\eta_t P\,\mathsf{RN}_{\delta_{\mathrm{num}}}(\widetilde B_t),
\qquad
W_{t+1}=W_t+\Delta W_t.
\label{eq:head-practical-rownorm}
\end{equation}
Under our assumptions, the momentum estimate is already centered in exact arithmetic, so the first projection only removes any accumulated numerical drift.
Row normalization can introduce a nonzero mean, which the final projection removes.
Every normalized row has norm at most one, and centering preserves pairwise differences, so $D(\Delta W_t)\leq2\eta_t$.
Consequently, $0\leq\eta_t\leq\varepsilon/(2R)$ ensures the required Hilbert distance bound for every $\norm{h}_2\leq R$.
The computation takes $O(Vd)$ time, uses one first-moment buffer, and implements each projection by subtracting the mean row.
Momentum and numerical stabilization preserve the diameter bound on the applied update.
The forward and backward estimates therefore remain applicable under their stated assumptions.

\section{Forward and backward perturbation bounds}
\label{app:perturbation-bounds}

A spectral constraint, as used by Muon, controls perturbations in both the forward and backward computations of a linear layer.
For a fixed input $h$ and incoming gradient $r$,
\[
\norm{\Delta Wh}_2\leq\norm{\Delta W}_{\mathrm{op}}\norm{h}_2,
\qquad
\norm{\Delta W^\top r}_2\leq\norm{\Delta W}_{\mathrm{op}}\norm{r}_2.
\]
We establish analogous bounds for our softmax modules below, comparing each module before and after an update at the same inputs.
Write $\tau=\tanh(\varepsilon/4)$.
The inequality relating Hilbert distance and total variation gives $\mathrm{TV}(p,q)\leq\tau$ whenever $\mathcal H(p,q)\leq\varepsilon$ \citep[Theorem~5.1]{cohen2024hyperboliccontractivityhilbertmetric}.

\subsection{The LM head}
\label{app:head-backward}

The diameter constraint controls the head's probability output and also bounds the change in the gradient passed to its hidden state.
Let $p=p_W(h)$ and $q=p_{W+\Delta W}(h)$ at a fixed input $\norm{h}_2\leq R$.
The forward bound is
\[
\mathcal H(p,q)\leq R D(\Delta W)\leq\varepsilon.
\]
For a target token $c$, denote the gradients passed to $h$ by $\lambda_h$ and $\lambda_h^+$.
Cross-entropy gives
\[
\lambda_h=W^\top(p-e_c),
\qquad
\lambda_h^+=(W+\Delta W)^\top(q-e_c),
\]
where $e_c$ is the target's one-hot vector.
Writing $w_j^\top$ and $\Delta w_j^\top$ for the rows of $W$ and $\Delta W$, subtraction gives
\[
\lambda_h^+-\lambda_h
=\sum_jq_j(\Delta w_j-\Delta w_c)+\sum_j(q_j-p_j)w_j.
\]

The first sum averages differences between update rows, so its norm is at most $D(\Delta W)$.
The second sum accounts for the redistributed probability.
Because $p$ and $q$ each have unit mass, the positive and negative coefficients $q_j-p_j$ have the same total mass, namely $\mathrm{TV}(p,q)$.
After normalization, they form two weighted averages of the head rows.
Their distance cannot exceed $D(W)$, giving
\[
\norm{\lambda_h^+-\lambda_h}_2
\leq D(\Delta W)+D(W)\mathrm{TV}(p,q)
\leq\frac{\varepsilon}{R}+\tau D(W).
\]
The current head appears through $D(W)$ because it transmits the changed probability gradient.

\subsection{Attention}
\label{app:attention-backward}

For attention, we again seek bounds on the change in the output distributions and the resulting change in the gradients passed to the hidden states.
The backward calculation now includes both the query and key uses of each hidden state.
We keep the inputs $\norm{h_\ell}_2\leq R$ and visible sets $\mathcal J_i$ fixed.
Let $M=W_QW_K^\top$, and let $\Delta M$ denote its change.
Equation~\ref{eq:attention-hilbert-spectral} gives
\[
\max_i\mathcal H(p_i,p_i^+)\leq\frac{2R^2}{\sqrt{d_h}}\norm{\Delta M}_{\mathrm{op}}\leq\varepsilon.
\]
The total variation bound recalled above then gives $\mathrm{TV}(p_i,p_i^+)\leq\tau$, where $\tau=\tanh(\varepsilon/4)$.

To connect this probability bound to backpropagation, let $r_i$ denote the incoming gradient on $p_i$.
We hold these incoming gradients fixed when comparing the module before and after the QK update.
Differentiating softmax gives the gradient with respect to the scores,
\[
a_{ij}=p_{ij}(r_{ij}-\mu_i),
\qquad
\mu_i=\sum_{j\in\mathcal J_i}p_{ij}r_{ij}.
\]
Each score $s_{ij}=h_i^\top M h_j/\sqrt{d_h}$ sends this gradient to both its query input and its key input.
Denote the corresponding contributions by $\lambda_i^{\mathrm Q}$ and $\lambda_{ij}^{\mathrm K}$:
\[
\lambda_i^{\mathrm Q}=\frac{M\sum_{j\in\mathcal J_i}a_{ij}h_j}{\sqrt{d_h}},
\qquad
\lambda_{ij}^{\mathrm K}=\frac{a_{ij}M^\top h_i}{\sqrt{d_h}}.
\]
The total gradient passed to $h_\ell$ adds its query contribution and its key contributions from all queries that can see it,
\[
\lambda_\ell=\lambda_\ell^{\mathrm Q}+\sum_{i:\ell\in\mathcal J_i}\lambda_{i\ell}^{\mathrm K}.
\]

There are two sources of change in these expressions: the QK product changes, and so does the score gradient.
Writing $\Delta a_i=a_i^+-a_i$ and $\Delta\lambda=\lambda^+-\lambda$, we obtain
\[
\begin{aligned}
\Delta\lambda_i^{\mathrm Q}
&=\frac{\Delta M\sum_{j\in\mathcal J_i}a_{ij}^+h_j+M\sum_{j\in\mathcal J_i}\Delta a_{ij}h_j}{\sqrt{d_h}},\\
\Delta\lambda_{ij}^{\mathrm K}
&=\frac{a_{ij}^+\Delta M^\top h_i+\Delta a_{ij}M^\top h_i}{\sqrt{d_h}}.
\end{aligned}
\]
Since every hidden state has norm at most $R$, the remaining quantities to bound are $\norm{a_i^+}_1$ and $\norm{\Delta a_i}_1$.
The first accompanies the product change, and the second accompanies the current product.

We establish these estimates for one row, temporarily omitting the index $i$.
Write $a_j(p)=p_j(r_j-\mu)$ with $\mu=\sum_jp_jr_j$.
Recall that $\osc(r)=\max_jr_j-\min_jr_j$.
For the size of $a(p)$, Cauchy--Schwarz reduces the calculation to the variance of the entries of $r$ under $p$.
With $m=(\max_jr_j+\min_jr_j)/2$ denoting the midpoint of their range,
\[
\norm{a(p)}_1^2
\leq\sum_jp_j(r_j-\mu)^2
=\sum_jp_j(r_j-m)^2-(\mu-m)^2
\leq\frac{\osc(r)^2}{4}.
\]

For the change in $a$, we use the probability redistribution measured by total variation.
Any change from $p$ to $q$ can be decomposed into transfers from coordinates that lose probability to coordinates that gain it.
The total mass transferred in this way is $\mathrm{TV}(p,q)$.
It therefore suffices to bound the effect of one transfer and add these bounds.

Suppose an amount $\alpha$ moves from coordinate $k$ to coordinate $j$, giving $q=p+\alpha(e_j-e_k)$.
Here $e_j,e_k$ are coordinate vectors and $0\leq\alpha\leq p_k$.
Subtracting the score gradients at fixed $r$ gives
\[
a(q)-a(p)=\alpha\bigl[(r_j-\mu)e_j-(r_k-\mu)e_k-(r_j-r_k)q\bigr].
\]
Because $q$ has unit mass,
\[
\norm{a(q)-a(p)}_1
\leq\alpha\bigl(|r_j-\mu|+|r_k-\mu|+|r_j-r_k|\bigr)
\leq2\alpha\,\osc(r).
\]
The last step uses the fact that the three absolute differences sum to twice the range of $r_j,r_k,\mu$.
All three values lie within the range of $r$, since $\mu$ is a weighted average of its entries.
Adding the bounds over the transfers proves $\norm{a(q)-a(p)}_1\leq2\osc(r)\mathrm{TV}(p,q)$.

Returning to the attention rows, the two estimates become
\[
\norm{a_i^+}_1\leq\frac{\osc(r_i)}2,
\qquad
\norm{\Delta a_i}_1\leq2\osc(r_i)\tau.
\]
Apply these to the query and key expressions above, using $\norm{h_\ell}_2\leq R$ and the triangle inequality:
\[
\begin{aligned}
\sum_\ell\norm{\Delta\lambda_\ell}_2
&\leq\frac{2R}{\sqrt{d_h}}\sum_i\left(\norm{\Delta M}_{\mathrm{op}}\norm{a_i^+}_1+\norm{M}_{\mathrm{op}}\norm{\Delta a_i}_1\right)\\
&\leq\frac{R}{\sqrt{d_h}}\left(\norm{\Delta M}_{\mathrm{op}}+4\norm{M}_{\mathrm{op}}\tau\right)\sum_i\osc(r_i).
\end{aligned}
\]
The factor of two in the first line accounts for the query and key contributions.
Both are included when a position attends to itself.
For fixed $M$ and incoming gradients, the bound tends to zero as $\varepsilon$ tends to zero under the stated spectral constraint.
For finite weight updates, $\Delta M$ includes the quadratic term $\Delta W_Q\Delta W_K^\top$.
The comparison concerns the QK probability module and does not include changes in the incoming gradients or the value pathway.

\section{Reproducing Figure~\ref{fig:rms-softmax-distributions}}
\label{app:figure-construction}

Figure~\ref{fig:rms-softmax-distributions} holds the initial distribution fixed and varies only the centered update.
We construct the initial logits by centering a quadratic profile over $i=1,\ldots,32$:
\[
a_i=-\left(\frac{i-16.5}{4}\right)^2,
\qquad
z_i=a_i-\frac{1}{32}\sum_{j=1}^{32}a_j,
\qquad
p=\softmax(z).
\]
The three updates below either shift a central block together or perturb individual preactivation coordinates within it.
Here, $\mathbf{1}_{\{\cdot\}}$ is the indicator function, equal to one when its condition holds and zero otherwise:
\[
\begin{aligned}
\textcolor{blue}{\Delta z_i^{\mathrm{blue}}}   &=2\mathbf{1}_{\{13\leq i\leq16\}}-2\mathbf{1}_{\{17\leq i\leq20\}},\\
\textcolor{violet}{\Delta z_i^{\mathrm{purple}}} &=2\mathbf{1}_{\{9\leq i\leq24\}}-1,\\
\textcolor{red}{\Delta z_i^{\mathrm{red}}}    &=2\mathbf{1}_{\{i=16\}}-2\mathbf{1}_{\{i=19\}}.
\end{aligned}
\]
Each update has zero mean and produces $q=\softmax(z+\Delta z)$.
Using
\[
\mathrm{RMS}(\Delta z)=\sqrt{\frac{1}{32}\sum_{i=1}^{32}(\Delta z_i)^2},
\qquad
\mathcal H(p,q)=\max_i\Delta z_i-\min_i\Delta z_i,
\]
we obtain the following values, together with the total variation $\mathrm{TV}(p,q)=\frac12\sum_i|p_i-q_i|$:
\[
\begin{array}{c|ccc}
 & \text{\textcolor{blue}{Blue}} & \text{\textcolor{violet}{Purple}} & \text{\textcolor{red}{Red}}\\ \hline
\mathrm{RMS}(\Delta z) & 1 & 1 & 1/2\\
\mathcal H(p,q) & 4 & 2 & 4\\
\mathrm{TV}(p,q) & \approx0.514 & \approx0.004 & \approx0.430
\end{array}.
\]
\textcolor{violet}{Purple} adds a common increment to coordinates $9$--$24$, which initially carry more than $99.5\%$ of the probability mass.
The pairwise odds among coordinates \(9,\ldots,24\) remain unchanged, and only a small amount of probability mass moves from the tails into this block.
\textcolor{blue}{Blue} and \textcolor{red}{red} instead alter these ratios, producing much larger redistribution under the same RMS bound.

\end{document}